\documentclass[11pt,letterpaper]{article}
\usepackage{texstyle}
\usepackage{times}
\usepackage{latexsym}
\usepackage{amsfonts}
\usepackage{amsmath}
\usepackage{amssymb}
\usepackage{multirow}
\usepackage{graphicx}

\emnlpfinalcopy

\def\emnlppaperid{***}

\newcommand\norm[1]{\left\lVert#1\right\rVert}

\title{Inducing Interpretability in Knowledge Graph Embeddings}

\author{Chandrahas \and Tathagata Sengupta \and Cibi Pragadeesh \and Partha Pratim Talukdar\\
  {\tt chandrahas@iisc.ac.in}}

\date{}

\begin{document}
\maketitle

\begin{abstract}
We study the problem of inducing interpretability in KG embeddings.
Specifically, we explore the Universal Schema \cite{ref:universalSchema} and propose a method to induce interpretability.
There have been many vector space models proposed for the problem,
however, most of these methods don't address the interpretability (semantics) of individual dimensions.
In this work, we study this problem and 
propose a method for inducing interpretability in KG embeddings using entity co-occurrence  statistics. 
The proposed method significantly improves the interpretability, while maintaining comparable performance in other KG tasks.

\end{abstract}

\section{Introduction}
Knowledge Graphs such as Freebase, WordNet etc. have become important resources for supporting many AI applications like web search, Q\&A etc. They store a collection of facts in the form of a graph. The nodes in the graph represent real world entities such as \textit{Roger Federer}, \textit{Tennis}, \textit{United States} etc while the edges represent relationships between them. 

These KGs have grown huge, but they are still not complete \cite{ref:toutanovaJE}. 
Hence the task of inferring new facts becomes important. 
Many vector space models have been proposed which can perform reasoning over KGs efficiently \cite{ref:transE}, \cite{ref:transH}, \cite{ref:transR}, \cite{ref:NTN}, \cite{ref:universalSchema}, \cite{ref:toutanovaJE} etc. 
These methods learn representations for entities and relations as vectors in a vector space, capturing global information about the KG. 
The task of KG inference is then defined as operations over these vectors. Some of these methods like \cite{ref:universalSchema}, \cite{ref:toutanovaJE} are capable of exploiting additional text data apart from the KG, resulting in better representations.


Although these methods have shown good performance in applications, they don't address the problem of understanding semantics of individual dimensions of the KG embedding.
A recent work \cite{ref:KSR} addressed the problem of learning semantic features for KGs. However, they don't directly use  vector space modeling.

In this work, we focus on incorporating interpretability in KG embeddings. Specifically, we aim to learn interpretable embeddings for KG entities by incorporating additional entity co-occurrence statistics from text data. This work is motivated by \cite{ref:teaLeaves} who presented automated methods for evaluating topics learned via topic modelling methods. We adapt these measures for the vector space model and propose a method to directly maximize them while learning KG embedding. To the best of our knowledge, this work presents the first regularization term which induces interpretability in KG embeddings.

\section{Related Work}
Several methods have been proposed for learning KG embeddings.
They differ on the modeling of entities and relations, usage of text data and interpretability of the learned embeddings.
We summarize some of these methods in following sections.


\subsection{Vector-space models for KG Embeddings}
A very effective and powerful set of models are based on translation vectors.
These models represent entities as vectors in $d$-dimensional space, $\mathbb{R}^d$ and relations as translation vectors from head entity to tail entity, in either same or a projected space.
TransE\cite{ref:transE} is one of the initial works,
which was later improved by many works [\cite{ref:transH}, \cite{ref:transR}, \cite{ref:transG}, \cite{ref:transA}, \cite{ref:transD}, \cite{ref:transM}]. 
Also, there are methods which are able to incorporate text data while learning KG embeddings.
\cite{ref:universalSchema} is one such method, which assumes a combined universal schema of relations from KG as well as text. 
\cite{ref:toutanovaJE} further improves the performance by sharing parameters among similar textual relations.


\subsection{Interpretability of Embedding}
While the vector space models perform well in many tasks, the semantics of learned representations are not directly clear.
This problem for word embeddings was addressed by \cite{ref:NNSE} where they proposed a set of constraints inducing interpretability.
However, its adaptation for KG embeddings hasn't been addressed.
A recent work \cite{ref:KSR} addressed a similar problem, where they learn coherent semantic features for entities and relations in KG.
Our method differs from theirs in the following two aspects.
Firstly, we use vector space modeling leading directly to KG embeddings while they need to infer KG embeddings from their probabilistic model.
Second, we incorporate additional information about entities which helps in learning interpretable embeddings.

\section{Proposed Method}

We are interested in inducing interpretability in KG embeddings and regularization is one good way to do it.
So we want to look at novel regularizers in KG embeddings.
Hence, we explore a measure of coherence proposed in \cite{ref:teaLeaves}.
This measure allows automated evaluation of the quality of topics learned by topic modeling methods by using additional Point-wise Mutual Information (PMI) for word pairs.
It was also shown to have high correlation with human evaluation of topics.

Based on this measure of coherence, we propose a regularization term. 
This term can be used with existing KG embedding methods (eg \cite{ref:universalSchema}) for inducing interpretability.
It is described in the following sections.

\subsection{Coherence}
\label{sec:coherenceSection}
In topic models, coherence of a topic can be determined by semantic relatedness among top entities within the topic.
This idea can also be used in vector space models by treating dimensions of the vector space as topics.
With this assumption, we can use a measure of coherence defined in following section for evaluating interpretability of the embeddings.
\subsubsection{$Coherence@k$}
\label{sec:coherence}
$Coherence@k$ has been shown to have high correlation with human interpretability of topics learned via various topic modeling methods\cite{ref:teaLeaves}.
Hence, we can expect interpretable embeddings by maximizing it.

Coherence for top $k$ entities along dimension $l$ is defined as follows:
\begin{equation}
Coherence@k^{(l)} = \sum_{i=2}^{k}\sum_{j=1}^{i-1}{p_{ij}}
\end{equation}
where $p_{ij}$ is PMI score between entities $e_i$ and $e_j$ extracted from text data.
$Coherence@k$ for the entity embedding matrix $\theta_e$ is defined as the average over all dimensions.
\begin{equation}
\label{eq:cohAtK}
Coherence@k = \frac{1}{d} \sum_{l=1}^{d} Coherence@k^{(l)}
\end{equation}

\subsubsection{Inducing coherence while learning embeddings}
\label{sec:coherenceDef}
We want to learn an embedding matrix $\theta_e$ which has high coherence (i.e. which maximizes $Coherence@k$).
Since $\theta_e$ changes during training, the set of top $k$ entities along each dimension varies over iterations.
Hence, directly maximizing $Coherence@k$ seems to be tricky.

An alternate approach could be to promote higher values for entity pairs having high PMI score $p_{ij}$.
This will result in an embedding matrix $\theta_e$ with a high value of $Coherence@k$ since high PMI entity pairs are more likely to be among top $k$ entities.

This idea can be captured by following coherence term
\begin{equation}
\mathcal{C}(\theta_e, P) = \sum_{i=2}^{n}\sum_{j=1}^{i-1} \norm { v(e_i)^\intercal v(e_j) - p_{ij} }^2
\end{equation}

where $P$ is entity-pair PMI matrix and $v(e)$ denote vector for entity $e$.
This term can be used in the objective function defined in Equation \ref{eq:objective}


\subsection{Entity Model (Model-E)}
\label{sec:modelE}
We use the Entity Model proposed in \cite{ref:universalSchema} for learning KG embeddings.
This model assumes a vector $v(e)$ for each entity and two vectors $v_s(r)$ and $v_o(r)$ for each relation of the KG. The score for the triple $(e_s, r, e_o)$ is given by,

\begin{equation}
\label{eq:modelE}
f(e_s, r, e_o) = v(e_s)^\intercal v_s(r) + v(e_o)^\intercal v_o(r)
\end{equation}

Training these vectors requires incorrect triples.
So, we use the closed world assumption.
For each triple $t \in \mathcal{T}$, we create two negative triples $t^-_o$ and $t^-_s$ by corrupting the object and subject of the triples respectively such that the corrupted triples don't appear in training, test or validation data.
The loss for a triple pair is defined as $loss(t, t^-) = - \log(\sigma(f(t) - f(t^-)))$.
Then, the aggregate loss function is defined as 

\begin{equation}
L(\theta_e, \theta_r, \mathcal{T}) = \frac{1}{|\mathcal{T}|}\sum_{t\in \mathcal{T}} \left(loss(t, t^-_o) + loss(t, t^-_s) \right)
\end{equation}

\subsection{Objective}
\label{sec:objective}
The overall loss function can be written as follows:
\begin{equation}
\label{eq:objective}
L(\theta_e, \theta_r, \mathcal{T}) + \lambda_c \mathcal{C}(\theta_e, P) + \lambda_r \mathcal{R}(\theta_e, \theta_r)
\end{equation}

Where $\mathcal{R}(\theta_e, \theta_r) = \frac{1}{2}\left(\norm{\theta_e}^2+\norm{\theta_r}^2\right)$ is the $L2$ regularization term and $\lambda_c$ and $\lambda_r$ are hyper-parameters controlling the trade-off among different terms in the objective function.
%

\section{Experiments and Results}

%

\subsection{Datasets}
We use the FB15k-237\cite{ref:FB15k_237} dataset for experiments.
It contains $14541$ entities and $237$ relations.
The triples are split into training, validation and test set having $272115$, $17535$ and $20466$ triples respectively.
For extracting entity co-occurrences, we use the textual relations used in \cite{ref:toutanovaJE}.
It contains around 3.7 millions textual triples, which we use for calculating PMI for entity pairs.

\subsection{Experimental Setup}
We use the method proposed in \cite{ref:universalSchema} as the baseline.
Please refer to Section \ref{sec:modelE} for more details.
For evaluating the learned embeddings, we test them on different tasks.
All the hyper-parameters are tuned using performance (MRR) on validation data.
We use 100 dimensions after cross validating among 50, 100 and 200 dimensions.
For regularization, we use $\lambda_r = 0.01$ (from $10,1,0.1,0.01$) and $\lambda_c = 0.01$ (from $10,1,0.1,0.01$) for $L2$ and coherence regularization respectively.
We use multiple random initializations sampled from a Gaussian distribution. 
For optimization, we use gradient descent and stop optimization when gradient becomes $0$ upto $3$ decimal places.
The final performance measures are reported for test data.

\subsection{Results}
\label{sec:expResults}
In following sections, we compare the performance of the proposed method with the baseline method in different tasks. Please refer to Table \ref{tab:results} for results.

%

\subsubsection{Interpretability}
For evaluating the interpretability,
we use $Coherence@k$ (Equation \ref{eq:cohAtK}) , automated and manual word intrusion tests.
In word intrusion test \cite{ref:teaLeavesChang}, top $k(=5)$ entities along a dimension are mixed with the bottom most entity (the intruder) in that dimension and shuffled.
Then multiple (3 in our case) human annotators are asked to find out the intruder. We use majority voting to finalize one intruder. Amazon Mechanical Turk was used for crowdsourcing the task and we used $25$ randomly selected dimensions for evaluation.
For automated word intrusion \cite{ref:teaLeaves}, we calculate following score for all $k+1$ entities

\begin{equation}
\label{eq:autoWI}
\text{AutoWI}(e_i) = \sum_{j=1, j\neq i}^{k+1}{p_{ij}}
\end{equation}

where $p_{ij}$ are the PMI scores.
The entity having least score is identified as the intruder.
We report the fraction of dimensions for which we were able to identify the intruder correctly.


As we can see in Table \ref{tab:results}, the proposed method achieves better values for $Coherence@5$ as a direct consequence of the regularization term, thereby maximizing coherence between appropriate entities. 
Performance on the word intrusion task also improves drastically as the intruder along each dimension is a lot easier to identify owing to the fact that the top entities for each dimension group together more
conspicuously.

\begin{table}[thb]
\centering
\small
\resizebox{\linewidth}{!}{%
\begin{tabular}{|c|c|l|c|c|l|l|}
\hline
Method     	& \multicolumn{6}{c|}{\textbf{Link Prediction}}                                     \\ \hline
                		& \multicolumn{2}{c|}{MRR} 			& \multicolumn{2}{c|}{MR} 			& \multicolumn{2}{l|}{Hits@10(\%)} 		\\ \hline
Baseline   		& \multicolumn{2}{c|}{$31.6 \pm 0.08$}   	& \multicolumn{2}{c|}{$121.9 \pm 1.80$}    & \multicolumn{2}{l|}{$48.3 \pm 0.39$}      	\\ \hline
Proposed  	& \multicolumn{2}{c|}{$30.4 \pm 0.08$}   	& \multicolumn{2}{c|}{$111.9 \pm 1.12$}    	& \multicolumn{2}{l|}{$46.8 \pm 0.08$}      	\\ \hline
                		& \multicolumn{6}{c|}{\textbf{Triple Classification}}                               \\ \hline
                		& \multicolumn{3}{c|}{AUC(\%)}              	& \multicolumn{3}{c|}{Accuracy(\%)}             		\\ \hline
Baseline        	& \multicolumn{3}{c|}{$72.9 \pm 0.16$} 	& \multicolumn{3}{c|}{$63.2 \pm 0.50$}                     	\\ \hline
Proposed  	& \multicolumn{3}{c|}{$73.2 \pm 0.28$}   	& \multicolumn{3}{c|}{$67.6 \pm 0.17$}                     	\\ \hline
                		& \multicolumn{6}{c|}{\textbf{Interpretability}}                                    \\ \hline
                		& \multicolumn{2}{c|}{AutoWI@5(\%)}        & \multicolumn{2}{c|}{Coherence@5}	& \multicolumn{2}{c|}{Manual WI(\%)}          		\\ \hline
Baseline        	& \multicolumn{2}{c|}{$6 \pm 4.14$}         	& \multicolumn{2}{c|}{$-47.4 \pm 4.68$}	& \multicolumn{2}{c|}{$12 $}                     \\ \hline
Proposed  	& \multicolumn{2}{c|}{$66 \pm 5.89$}         & \multicolumn{2}{c|}{$-12.5 \pm 4.48$} 	& \multicolumn{2}{c|}{$84$}                    \\ \hline
\end{tabular}
}
\caption{Results on test data. The proposed method significantly improves interpretability while maintaining comparable performance on KG tasks (Section \ref{sec:expResults}).}
\label{tab:results}
\end{table}

\begin{table}[thb]
\centering
\resizebox{\linewidth}{!}{%
\begin{tabular}{|l|}
\hline

\multicolumn{1}{|c|}{\textbf{Top 5}} \\ \hline
\multicolumn{1}{|c|}{\textbf{Baseline}}   \\ \hline
 -\textbf{Jurist}, \textbf{Pipe organ}, USA, \textbf{Lions Gate Entertainment}, UK \\
-Guitar, \textbf{71st Academy Awards}, \textbf{Jurist}, Piano, Bass guitar \\ 
-Actor, \textbf{Official Website}, Screenwriter, Film Producer, \textbf{USA}\\ 
-\textbf{Jurist}, USA, \textbf{Marriage}, \textbf{Male}, UK \\ 
-\textbf{Pipe organ}, \textbf{Official Website}, Actor, Film Producer, Screenwriter  \\ \hline

\multicolumn{1}{|c|}{\textbf{Proposed Method}} \\ \hline
-Juris Doctor, Business Administration, Biology, Psychology, BS \\ 
-Bachelor of Arts, PhD, Bachelor's degree, BS, MS\\  
-European Union, Europe, Netherlands, Portugal, \textbf{Government} \\
-UK, Hollywood, \textbf{DVD}, London, Europe\\ 
-Hollywood, Academy Awards, \textbf{USA}, DVD, \textbf{Los Angeles}\\ \hline

\end{tabular}
}
\caption[results2]{Top 5 and bottom most entities for randomly selected dimensions. As we see, the proposed method produces more coherent entities compared to the baseline. Incoherent entities are marked in bold face. \footnotemark}
\label{tab:top5Ents}
\end{table}

\subsubsection{Link Prediction}
In this experiment, we test the model's ability to predict the best object entity for a given subject entity and relation.
For each of the triples, we fix the subject and the relation and rank all entities (within same category as true object entity) based on their score according to Equation \ref{eq:modelE}.
We report Mean Rank (MR) and Mean Reciprocal rank (MRR) of the true object entity and Hits@10 (the number of times true object entity is ranked in top 10) as percentage. 
The objective of the coherence regularization term being tangential to that of the original loss function, is not expected to affect performance on the link prediction task. However, the results show a trivial drop of $1.2$ in MRR as the coherence term gives credibility to triples that are otherwise deemed incorrect by the closed world assumption. 

\footnotetext{We have used abbreviations for BS (Bachelor of Science), MS (Master of Science), UK (United Kingdom) and USA (United States of America). They appear as full form in the data.}

\subsubsection{Triple Classification}
In this experiment, we test the model on classifying correct and incorrect triples.
For finding incorrect triples, we corrupt the object entity with a randomly selected entity within the same category.
For classification, we  use validation data to find the best threshold for each relation by training an SVM classifier and later use this threshold for classifying test triples. We report the mean accuracy and mean AUC over all relations.



We observe that the proposed method achieves slightly better performance for triple classification improving the accuracy by $4.4$.
The PMI information adds more evidence to the correct triples which are related in text data, generating a better threshold that more accurately distinguishes correct and incorrect triples. 

\subsection{Qualitative Analysis of Results}
Since our aim is to induce interpretability in representations, in this section, we evaluate the embeddings learned by the baseline as well as the proposed method.
For both methods, we select some dimensions randomly and present top 5 entities along those dimensions. The results are presented in Table \ref{tab:top5Ents}.

As we can see from the results, the proposed method produces more coherent entities than the baseline method.

%
%


\section{Conclusion and Future Works}
In this work, we proposed a method for inducing interpretability in KG embeddings using a coherence regularization term.
We evaluated the proposed and the baseline method on the interpretability of the learned embeddings.
We also evaluated the methods on different KG tasks and compared their performance.
We found that the proposed method achieves better interpretability while maintaining comparable performance on KG tasks.
As next steps, we plan to evaluate the generalizability of the method with more recent KG embeddings.


\bibliography{myrefs}
\bibliographystyle{bibstyle}

\end{document}